%% file: main.tex
\documentclass[runningheads]{llncs}

\usepackage{eccv}

\usepackage{eccvabbrv}

\usepackage{graphicx}
\usepackage{booktabs}

\usepackage[accsupp]{axessibility}  

\usepackage{hyperref}

\usepackage{orcidlink}

\usepackage{fontawesome5}
\usepackage{tikz}
\usetikzlibrary{positioning, fit, backgrounds, shapes, shadows, arrows.meta, calc}

\usepackage{multirow}

\begin{document}

\title{TransHands: Repurposing Human Pose Encoders as Hand Pose Encoders} 

\titlerunning{TransHands: Repurposing Human Pose Encoders as Hand Pose Encoders}

\author{
Milo Piccioli\inst{1} \and
Gianluca Amprimo\inst{1}\orcidlink{0000-0003-4061-8211} \and
Claudia Ferraris\inst{2}\orcidlink{0000-0001-5381-4794} \and
Gabriella Olmo\inst{1}\orcidlink{0000-0002-3670-9412}}

 \authorrunning{M.Piccioli et al.}

\institute{Politecnico di Torino, Torino, Italy \email{s359009@studenti.polito.it, \{gianluca.amprimo, gabriella.olmo\}@polito.it} \and
Consiglio Nazionale delle Ricerche, Torino, Italy
\email{claudia.ferraris@cnr.it}\\}

\maketitle

\begin{abstract}
  Lifting 3D hand poses from 2D monocular representations remains challenging due to the limited availability of large-scale, diverse 3D-annotated hand datasets, in contrast to the abundance of human body motion data. We address this limitation by transferring motion representations learned from large body pose corpora to the hand domain.
  We introduce \textbf{TransHands}, a backbone-agnostic transfer learning framework that enables pre-trained human motion encoders to be effectively adapted for 3D hand pose estimation from 2D pose inputs. Rather than training hand-specific biomechanical models from scratch, TransHands combines a two-stage training and fine-tuning strategy with a lightweight hand-specific input adaptation module that aligns hand kinematics with the representation space learned for full-body motion.
  We evaluate TransHands across four state-of-the-art motion modeling architectures, including transformer-based, graph-based, and frequency-domain models. Results demonstrate that motion priors learned from body pose data transfer consistently across architectures, yielding consistent accuracy gains, strong cross-domain generalization, particularly in challenging egocentric settings, and applicability for downstream tasks in real-world contexts.
\end{abstract}

\input{src/sec_introduction}

\input{src/sec_relatedWorks}

\input{src/sec_method}

\input{src/sec_experiments}

\input{src/sec_robustness}

\input{src/sec_conclusion}





%
%
\bibliographystyle{splncs04}
\bibliography{references}
\end{document}

%% file: src/sec_introduction.tex
\begin{figure}[t]
\centering
\includegraphics[width=0.8\textwidth]{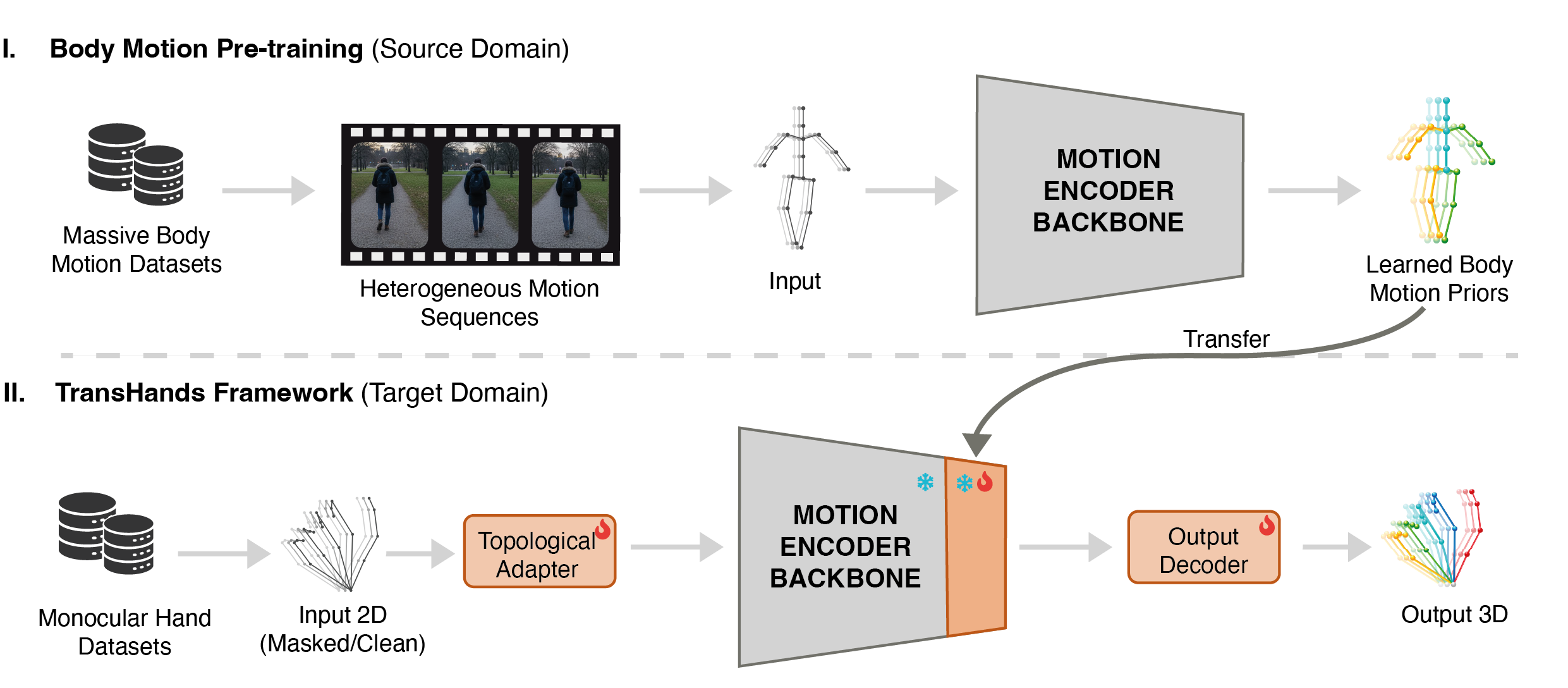}
\caption{\textbf{Framework Overview.}
TransHands bridges the domain gap between body and hand motion through a backbone-agnostic transfer learning strategy.
We leverage \textbf{(I) Body Motion Pre-training} to capture spatio-temporal priors from heterogeneous body datasets. 
These priors are repurposed for hand pose estimation via \textbf{(II) TransHands Framework}, using a learnable \textit{Topological Adapter} in input to align monocular hand sequences with the pre-trained body manifold.}
\label{fig:framework}
\end{figure}

\section{Introduction}
\label{sec:intro}

Estimating 3D hand pose from 2D joint coordinates (\textit{2D-to-3D lifting}) is a fundamental challenge in computer vision. While significant progress has been made in full-body pose estimation, benefiting from large-scale motion capture datasets such as Human3.6M~\cite{ionescu2014human36m} and AMASS~\cite{mahmood2019amassarchivemotioncapture}, hand pose estimation remains constrained by the limited availability of diverse, large-scale annotated hand datasets. This data scarcity poses a critical bottleneck: hands exhibit complex biomechanical constraints and high-frequency articulations that are difficult to capture and annotate at scale. Existing hand-specific datasets \cite{zimmermann2019freihand, fan2023arcticdatasetdexterousbimanual} are orders of magnitude smaller than their body pose counterparts, limiting the capacity of data-driven models to learn robust kinematic priors. Meanwhile, state-of-the-art (SOTA) human motion encoders~\cite{zhu2023motionbertunifiedperspectivelearning, zhao2023poseformerv2exploringfrequencydomain, zhang2022mixsteseq2seqmixedspatiotemporal} have demonstrated remarkable proficiency in modeling spatio-temporal dynamics from abundant body motion corpora.

We pose a natural question: \textit{Can motion priors learned from large-scale human body datasets be transferred to model hand biomechanical knowledge?} Despite the topological differences, both domains share fundamental biomechanical principles such as temporal smoothness, joint angle constraints, and hierarchical structure. Recent work has demonstrated the effectiveness of pre-training, transfer learning and cross-domain adaption in pose estimation~\cite{zhu2023motionbertunifiedperspectivelearning, li2025binaryhpe3dhumanpose, zhao2025analyzingsynthetictorealdomaingap}, suggesting that learned motion representations can generalize across modalities; however, a systematic study on the transfer of body motion priors to the hand domain is still lacking.

To this end, we introduce \textbf{TransHands}, a backbone-agnostic framework that enables pre-trained human motion encoders to be effectively repurposed for 3D hand pose estimation from 2D inputs. Our key insight is that the high-level temporal dynamics and biomechanical knowledge encoded in body pose models can be preserved and adapted to hand kinematics through \textit{learnable topological alignment} rather than architectural redesign. TransHands achieves this by encapsulating frozen body encoders within lightweight adaptation modules that bridge the domain gap while preserving the original kinematic reasoning capabilities. As illustrated in Figure \ref{fig:framework}, our approach comprises three main components: (1)~a \textbf{Topological Adapter} in input that aligns hand joint hierarchies with the latent space of body-centric encoders, (2)~a pre-trained \textbf{motion backbone} on large body datasets that extracts spatio-temporal features, and (3)~a \textbf{task-specific decoder} that reconstructs 3D hand coordinates.

We validate TransHands across four diverse motion encoding architectures spanning transformer-based~\cite{ionescu2014human36m, zhang2022mixsteseq2seqmixedspatiotemporal}, frequency-domain~\cite{zhao2023poseformerv2exploringfrequencydomain}, and graph-based~\cite{yan2018spatialtemporalgraphconvolutional} models. Our experiments demonstrate that: (1)~motion priors transfer consistently across architectural families, yielding consistent accuracy gains, (2)~the learned representations exhibit strong cross-domain generalization, particularly in challenging egocentric settings, and (3)~the framework requires minimal modification to existing backbones, enabling rapid integration of future motion modeling advances.

In summary, our primary contributions are threefold: 
\begin{itemize}
    \item To the best of our knowledge, we present the first systematic study on transferring body motion priors to hand pose estimation across multiple SOTA architectures.
    \item We propose a modular framework that decouples topological adaptation from temporal modeling, enabling backbone-agnostic transfer.
    \item We demonstrate that biomechanical knowledge learned from body datasets provides strong inductive biases for modeling hand kinematics, opening new directions for data-efficient hand pose estimation and downstream applications in gesture recognition.
\end{itemize}

%% file: src/sec_relatedWorks.tex
\section{Related Works}
\label{sec:related}
Early learning-based approaches formulated 2D-to-3D pose uplifting as a direct regression, showing that even simple feed-forward models can exploit geometric consistency and temporal smoothness when trained with paired 2D–3D data \cite{martinez2017simple,drover2018can}.
The emergence of large-scale human motion datasets enabled a shift toward data-driven learning of kinematic priors. Datasets such as AMASS~\cite{mahmood2019amassarchivemotioncapture} facilitated the pretraining of spatio-temporal motion encoders that implicitly capture biomechanical properties, including joint coordination and long-range temporal dependencies. Building on this paradigm, recent work has demonstrated the effectiveness of powerful sequence models for 2D-to-3D lifting, including spatio-temporal transformers \cite{zhu2023motionbertunifiedperspectivelearning,zhang2022mixsteseq2seqmixedspatiotemporal}, frequency-domain formulations \cite{zhao2023poseformerv2exploringfrequencydomain}, and graph-based or hybrid architectures leveraging skeletal structure as an inductive bias \cite{yan2018spatialtemporalgraphconvolutional, mehraban2023motionagformerenhancing3dhuman}. These approaches encode biomechanical knowledge implicitly through learned motion dynamics rather than explicit constraints.
In parallel, biomechanical plausibility has been addressed through model-based lifting approaches that fit parametric human models to monocular observations. Statistical body models such as SMPL \cite{loper2023smpl} and GHUM \cite{xu2020ghum} provide strong pose and shape priors and are widely used for monocular 3D reconstruction and tracking, including real-time systems such as BlazePose GHUM Holistic \cite{grishchenko2022blazepose}. For hands, analogous constraints are commonly imposed through the MANO statistical model \cite{romero2017embodied}, to ensure anatomical validity of estimated coordinates \cite{zimmermann2019freihand, fan2023arcticdatasetdexterousbimanual}. While effective, these model-based methods rely on carefully engineered optimization pipelines and explicit skeletal modeling, limiting flexibility and cross-domain adaptability.
Despite the success of data-driven motion encoders for full-body lifting, hand pose estimation remains comparatively data-limited. Transfer learning offers a natural alternative: pre-training on large body datasets followed by downstream adaptation has shown consistent benefits in tasks related to pose estimation \cite{shan2022pstmo, zhu2023motionbertunifiedperspectivelearning}. To date, these approaches have largely assumed a shared skeletal topology, leaving open whether motion encoders pre-trained for body pose lifting can be systematically repurposed for hand pose estimation despite substantial topological mismatch. Our work addresses this gap by reframing hand pose estimation as a biomechanical lifting problem and adapting 2D-to-3D body motion encoders to hand kinematics via lightweight, modular alignment modules.

%% file: src/sec_method.tex
\section{Method}
\label{sec:method}

\subsection{Problem Formulation}
\label{subsec:problem}

We formulate 3D hand pose estimation as a sequence-to-sequence lifting task. Specifically, we focus on the geometric lifting process, assuming the availability of 2D skeletal coordinates either from off-the-shelf detectors or ground-truth projections. This allows us to isolate the motion transfer capabilities from the noise inherent in direct 2D-to-3D estimation from RGB sources.
Given a sequence of input coordinates $X \in \mathbb{R}^{T \times J_H \times C_{in}}$, where $T$ denotes the sequence length, $J_H=21$ represents the hand joints according to the MANO~\cite{romero2017embodied} topology, and $C_{in} \in \{2, 3\}$ is the backbone-dependent input dimensionality, our objective is to regress the trajectory of 3D coordinates $Y \in \mathbb{R}^{T \times J_H \times 3}$.

\input{fig/fig_architecture}

A fundamental challenge in leveraging body motion priors for this task is the \textit{topological mismatch}: SOTA body encoders are typically optimized on skeletal full pose structures, rendering them incompatible with hand geometry.
To bridge this domain gap without retraining the kinematic engine from scratch, we propose \textbf{TransHands}, the modular framework illustrated in Figure~\ref{fig:arch}.
It encapsulates a frozen body encoder $\mathcal{E}_{\text{body}}$ within learnable adaptation layers. The mapping function is formalized as:
\begin{equation}
\label{eq:transhand_forward}
    \hat{Y} = \mathcal{D}_{\text{out}}\left( \mathcal{P}\left( \mathcal{E}_{\text{body}}\left( \mathcal{A}_{\text{in}}(X) \right) \right) \right)
\end{equation}
Here, $\mathcal{A}_{\text{in}}$ projects the hand topology into the body's latent kinematic space, $\mathcal{P}$ aligns the extracted features via projection,
and $\mathcal{D}_{\text{out}}$ decodes the processed representations back to the target $\mathbb{R}^{3}$ hand manifold. This formulation effectively decouples the learning of temporal dynamics, which is handled by the pre-trained $\mathcal{E}_{\text{body}}$, from the spatial adaptation required for the new topology.

\subsection{Topological Adapter}
\label{subsec:input_adapter}

The primary function of the Topological Adapter ($\mathcal{A}_{in}$) is to align hand kinematics with the representation space learned for full-body motion, resolving the dimensional and topological discrepancies between hand joints and body-centric pre-trained encoders. While SOTA body pose estimators are typically optimized on skeletal structures with $J_B=17$ joints, as in Human3.6M~\cite{ionescu2014human36m} topology or $J_B=18$, as in ST-GCN~\cite{yan2018spatialtemporalgraphconvolutional}, standard hand models such as MANO define a distinct topology consisting of $J_H=21$ joints. Direct application of the frozen body encoder $\mathcal{E}_{\text{body}}$ is therefore infeasible due to this mismatch ($J_H \neq J_B$). Moreover, the two graphs share no common joints except for the wrist, which, however, plays significantly different roles in the two kinematic graphs (i.e., the root node in the hand graph vs. a distal node in the body pose graph).

To address this, we formulate the topological transformation as a continuous dynamic process using a Neural Ordinary Differential Equation (Neural ODE)~\cite{chen2019neuralordinarydifferentialequations}. Given the flattened input coordinates $x \in \mathbb{R}^{J_H \cdot C_{in}}$, an initial encoder maps them into a latent state $z(0) \in \mathbb{R}^{d_{emb}}$, where $d_{emb}$ is the latent dimension. The continuous transformation of the hidden state is then defined by the ODE:
\begin{equation}
    \frac{dz(t)}{dt} = f_\theta(z(t), t)
\end{equation}
where $f_\theta$ is parameterized by a continuous dynamics neural network equipped with time-embeddings. We solve the initial value problem using a fixed-step fourth-order Runge--Kutta (RK4) integrator over the integration interval $t \in [0,1]$ to obtain the terminal state $z(1)$. Finally, a decoder maps $z(1)$ into the compatibility vector $z_{body} \in \mathbb{R}^{J_B \cdot C_{enc}}$ required by the backbone. This continuous flow formulation allows for a highly non-linear topological alignment without suffering from the rigidity of discrete linear projections. 

Critically, $\mathcal{A}_{in}$ acts as a \textbf{dynamic interface} that abstracts the underlying backbone requirements from the raw input data:
\begin{itemize}
    \item \textbf{Channel Alignment:} The adapter adjusts the channel dimensionality $C_{enc}$ to match the specific requirement of the selected backbone. For instance, it synthesizes a confidence channel ($C=3$) for MotionBERT~\cite{zhu2023motionbertunifiedperspectivelearning}, while preserving raw coordinates ($C=2$) for architectures like MixSTE~\cite{zhang2022mixsteseq2seqmixedspatiotemporal} and PoseFormerV2~\cite{zhao2023poseformerv2exploringfrequencydomain}.
    \item \textbf{Latent Manifold Alignment:} By optimizing the ODE trajectory, the adapter smoothly projects the hand topology onto the body's latent kinematic manifold, effectively reusing pre-trained priors without structural modifications to the frozen $\mathcal{E}_{\text{body}}$.
\end{itemize}

\subsection{Motion Encoder Backbones}
\label{subsec:motion_engines}

The core component of our framework is the kinematic encoder $\mathcal{E}_{\text{body}}$, which processes the adapted sequence $Z_{\text{body}} = \mathcal{A}_{in}(X)$ to extract spatio-temporal features. To enforce the transfer of high-level motion priors (e.g., velocity continuity, biomechanical constraints), we adopt a \textit{Partial Unfreezing} strategy.
Instead of fine-tuning, we strategically enable gradients only for the deepest layers of the backbone, together with its embedding and output layers. This design choice ensures that the model reuses learned body dynamics while allowing the final representation stages to adapt to hand-specific micro-articulations.
We validate the versatility of our approach by integrating four distinct SOTA architectures serving as Motion Encoder Backbone.

\textbf{Spatio-Temporal Transformers.} We integrate \textbf{MotionBERT}~\cite{zhu2023motionbertunifiedperspectivelearning}, utilizing its Dual-Stream Spatio-Temporal Transformer (DSTformer) to capture long-range dependencies. Additionally, we integrate \textbf{MixSTE}~\cite{zhang2022mixsteseq2seqmixedspatiotemporal}, a Seq2Seq model that decouples spatial and temporal attention, allowing us to evaluate the benefits of fine-grained frame-to-frame modeling for hand micro-articulations.

\textbf{Frequency-Domain Transformers.} To address long-sequence efficiency, we incorporate \textbf{PoseFormerV2}~\cite{zhao2023poseformerv2exploringfrequencydomain}, which operates in the frequency domain via Discrete Cosine Transform (DCT). This architecture is particularly relevant for analyzing the spectral properties of rapid hand gestures.

\textbf{Graph-Based Architectures.} To benchmark against purely convolutional approaches, we integrate the standalone implementation of \textbf{ST-GCN}~\cite{yan2018spatialtemporalgraphconvolutional}, demonstrating the framework's adaptability to non-transformer backbones. For this backbone the transfer is only partial: the temporal convolutions are initialized from the Kinetics checkpoint, whereas the graph branch (spatial convolutions, edge importance, and adaptive adjacency) is trained from scratch.

\subsection{Latent Space Alignment}
\label{subsec:projection}

Since different backbones produce feature maps with varying distributions and sequential dynamics (covariate shift), we introduce a projection module $\mathcal{P}$ to align the encoder output $f_{enc}$ with the decoding task. Instead of a standard linear bottleneck or attention-based transformer, we utilize a multi-scale retention mechanism~\cite{sun2023retentivenetworksuccessortransformer}. Formally:
\begin{equation}
    f_{proj} = W_p \cdot \text{LayerNorm}(f_{enc}) 
\end{equation}
Subsequently, the sequence is processed by a stack of retention layers with residual connections:
\begin{equation}
    \tilde{f} = f_{proj} + \text{Retention}(\text{LayerNorm}(f_{proj})), \quad f_{ret} = \tilde{f} + \text{FFN}(\tilde{f})
\end{equation}
where the retention operator applies learned exponential temporal decay to the position weights. The projection head standardizes heterogeneous encoder outputs to a common 256-dimensional latent before the decoding stage, preventing feature distortion during fine-tuning~\cite{kumar2022finetuningdistortpretrainedfeatures}.

\subsection{Task-Specific Output Decoder}
\label{subsec:decoder}

The reconstruction of the target 3D pose is performed by the module $\mathcal{D}_{out}$, a task-specific regressor. We design this decoder as a deep Multilayer Perceptron (MLP) utilizing Layer Normalization~\cite{ba2016layernormalization} for stable training across sequences. The architecture follows: 
\begin{gather}
    h_1 = \text{ReLU}(\text{LN}(W_1 f_{ret} + b_1)), \quad
    h_2 = \text{ReLU}(\text{LN}(W_2 h_1 + b_2)), \\
    \hat{Y} = \text{Reshape}_{J_H \times 3}(W_3 h_2 + b_3)
\end{gather}
where $W_1 \in \mathbb{R}^{512 \times 256}$, $W_2 \in \mathbb{R}^{512 \times 512}$, and $W_3 \in \mathbb{R}^{63 \times 512}$ (for $J_H=21$ joints).
The inclusion of Layer Normalization combined with sequential hidden expansion is critical to prevent training instabilities when decoding the narrow bottleneck. The depth of this module (three linear transformations mapping $256 \to 512 \to 512 \to 63$) is essential to model the non-linear inverse kinematics required to recover complex finger articulations from the compressed latent representation.

%% file: fig/fig_architecture.tex
\begin{figure*}[t]
\centering

\definecolor{paperBlue}{RGB}{227, 242, 253}
\definecolor{orange}{RGB}{241, 200, 173}
\definecolor{orangeDark}{RGB}{189, 87, 23}
\definecolor{paperDarkBlue}{RGB}{21, 101, 192}
\definecolor{paperGray}{RGB}{245, 245, 245}
\definecolor{paperDarkGray}{RGB}{97, 97, 97}
\definecolor{fireRed}{RGB}{229, 57, 53}
\definecolor{iceCyan}{RGB}{0, 188, 212}

\resizebox{1.0\textwidth}{!}{
\begin{tikzpicture}[
    font={\small}, 
    >=stealth,
    icon/.style={ anchor=north east, inner sep=2pt, font=\scriptsize },
    learnable/.style={
        rectangle, rounded corners=2pt, 
        draw=orangeDark!80, fill=orange, thick, 
        minimum height=1.3cm, minimum width=2.4cm, 
        align=center, anchor=center,
        drop shadow={opacity=0.1, shadow xshift=1mm, shadow yshift=-1mm}
    },
    miniFrozen/.style={
        rectangle, draw=paperDarkGray!50, fill=paperGray, 
        minimum height=0.7cm, minimum width=0.45cm, anchor=center
    },
    imgNode/.style={ inner sep=0pt, anchor=center }, 
    engineBox/.style={
        rectangle, rounded corners=3mm, 
        draw=paperDarkGray!40, 
        dashed, thick,         
        fill=gray!3, anchor=center
    },
    legendBox/.style={
        rectangle, rounded corners=2pt, draw=gray!30, fill=white, 
        font=\scriptsize, align=left, inner sep=4pt,
        drop shadow={opacity=0.05}
    },
    tensorLabel/.style={
        font=\tiny, scale=0.9, text=gray!90, midway, above=2pt
    }
]
    \node[imgNode] (input) at (0,0) {
        \includegraphics[width=2.3cm]{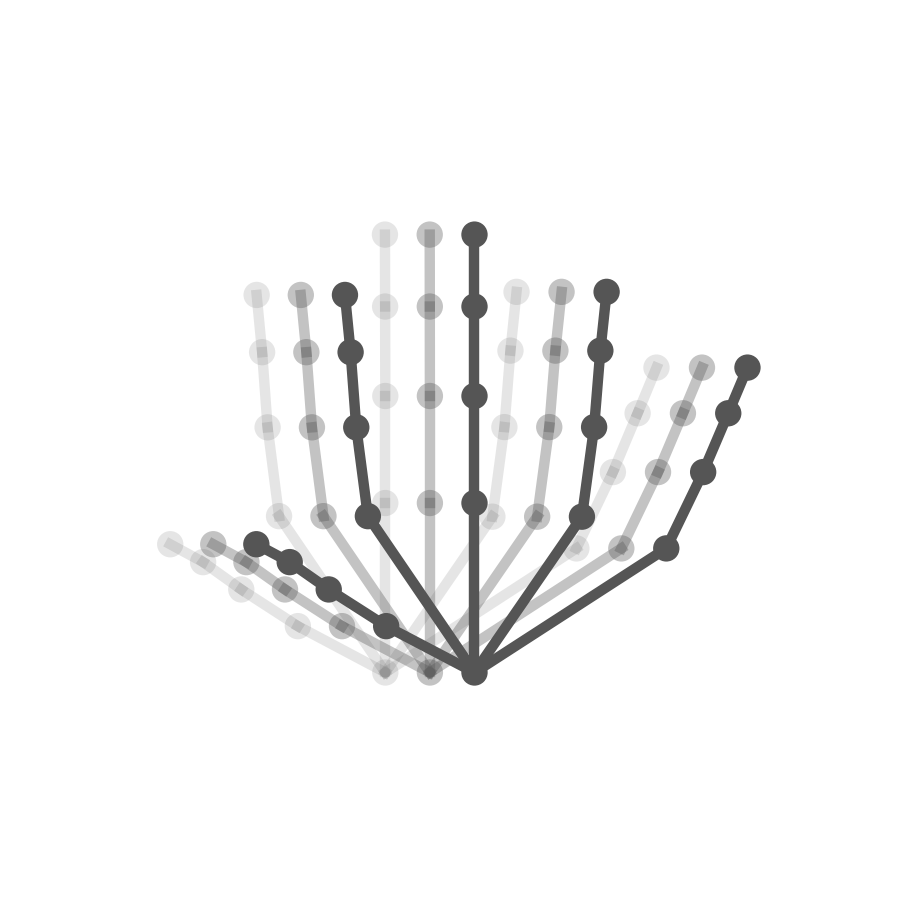}
    };
    \node[below=0.8cm of input.center, font={\footnotesize}, text=gray] {Input ($X$)};

    \node[learnable] (adapter) at (2.5,0) {
        Topological\\ \textbf{Adapter}\\
        \scriptsize $\mathcal{A}_{in}$
    };
    \node[icon, text=fireRed] at (adapter.north east) {\faFire};

    
    \node[miniFrozen] (b1) at (6.5,0) {}; 
    \node[miniFrozen] (b2) at (7.2,0) {};
    \node[font=\footnotesize, text=gray!60] (dots) at (8.1,0) {$\cdots \cdots$};
    \node[miniFrozen] (b3) at (9.0,0) {};
    \node[miniFrozen] (last) at (9.8,0) {};
    
    \node[iceCyan, above=0.3cm of b1.center, font=\tiny] {\faSnowflake};
    \node[iceCyan, above=0.3cm of b2.center, font=\tiny] {\faSnowflake};
    \node[iceCyan, above=0.3cm of b3.center, font=\tiny] {\faSnowflake};
    \node[iceCyan, xshift=-3pt, above=0.3cm of last.center, font=\tiny] {\faSnowflake};
    \node[fireRed, xshift=3pt, above=0.3cm of last.center, font=\tiny] {\faFire};

    \node[font={\footnotesize}, text=paperDarkGray, above=0.8cm of dots.center] 
        (label_center) {\textsc{Motion Encoder Backbone}};

    \begin{scope}[on background layer]
        \node[engineBox, fit=(b1)(last)(label_center), inner sep=12pt] (engine_container) {};
    \end{scope}

    \node[learnable] (decoder) at (13.5,0) {
        Output\\ \textbf{Decoder}\\
        \scriptsize $\mathcal{P} \rightarrow \mathcal{D}_{out}$
    };
    \node[icon, text=fireRed] at (decoder.north east) {\faFire};

    \node[imgNode] (output) at (16.5,0) {
        \includegraphics[width=2.3cm]{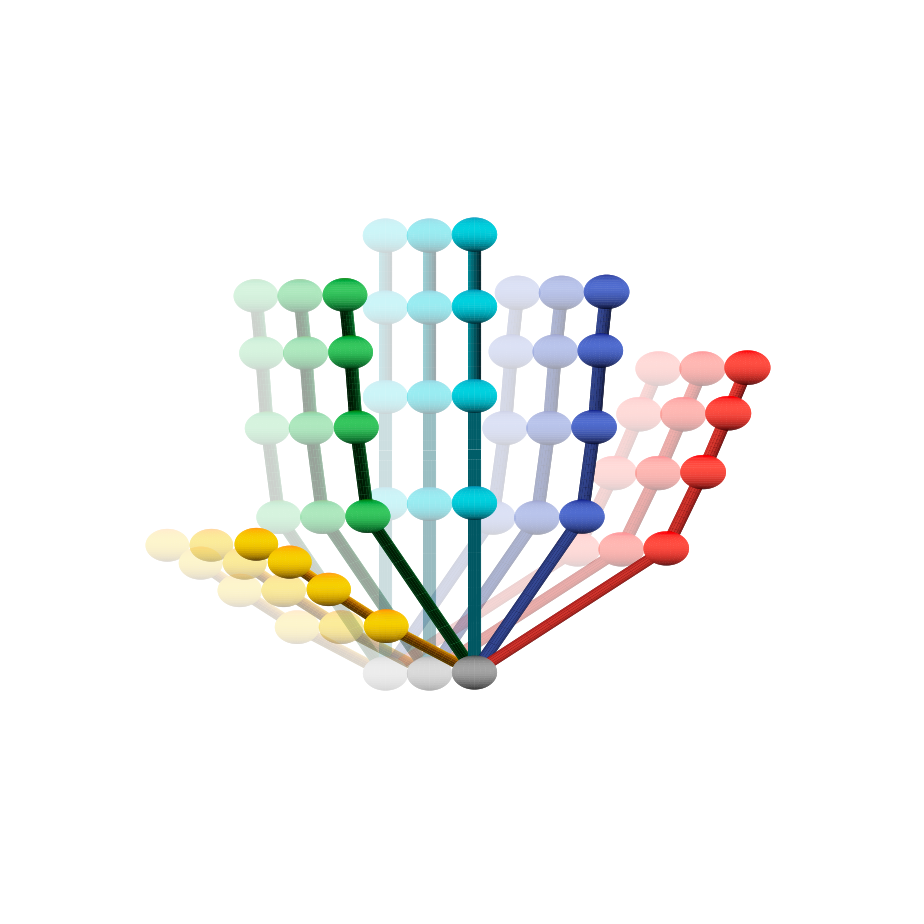}
    };
    \node[below=0.8cm of output.center, font={\footnotesize}, text=gray] {Output ($Y$)};

    
    \draw[->, thick, gray!80] ([xshift=-10pt]input.east) -- (adapter.west);
    \draw[->, thick, gray!80] (adapter.east) -- node[tensorLabel] {mapped ($J_{B}$)} (engine_container.west |- adapter.center);
    \draw[->, thick, gray!80] (engine_container.east |- decoder.center) -- node[tensorLabel] {features ($J_{B}$)} (decoder.west);
    \draw[->, thick, gray!80] (decoder.east) -- node[tensorLabel] {($J_{H}$)} ([xshift=10pt]output.west);

    \node[draw=gray!40, fill=white, dashed, rounded corners=2pt, below=0.8cm of engine_container, align=center, font={\footnotesize}, inner sep=6pt] (list) {
        \textbf{Plug-and-Play Backbones:}\\
        \textit{MotionBERT} $\cdot$ \textit{MixSTE} $\cdot$ \textit{PoseFormerV2} $\cdot$ \textit{ST-GCN}
    };
    \draw[dashed, thin, gray, ->] (list.north) -- (engine_container.south);

    \node[legendBox, anchor=south east] at (output.north east |- label_center.north) {
        \faFire \, Trainable \quad 
        \faSnowflake \, Frozen \
    };

\end{tikzpicture}
}
\caption{\textbf{Overview of the TransHands Framework.} To bridge the domain gap between body and hand topology, we employ a learnable \textbf{Topological Adapter} ($\mathcal{A}_{in}$) in input that aligns hand joints with the latent space of a pre-trained \textbf{Motion Encoder Backbone}. 
The architecture adopts a \textit{Partial Unfreezing} strategy: initially, the entire backbone is frozen (\faSnowflake) to leverage pre-trained human motion priors (Stage 1); subsequently, the final block is unfrozen (\faFire) to specialize the features for hand-specific kinematics (Stage 2).
The Output Decoder then maps these refined features to the final 3D hand coordinates.}
\label{fig:arch}
\end{figure*}

%% file: src/sec_experiments.tex
\section{Experiments}
\label{sec:experiments}

\subsection{Setup}
\label{subsec:setup}

\textbf{Datasets and Data Rationale.} 
To validate the effectiveness of TransHands, we adopt \textbf{Re:InterHand}~\cite{reinterhand} as our primary benchmark, a large-scale dataset of relighted 3D interacting hands with high-fidelity MANO-based annotations captured in a controlled multi-view studio setting.
To assess robustness across diverse scenarios, we employ \textbf{AssemblyHands}~\cite{ohkawa2023assemblyhands} (egocentric) and \textbf{GigaHands}~\cite{fu2025gigahandsmassiveannotateddataset} (bimanual hand activities). These datasets provide high-fidelity 2D ground-truth alongside their 3D annotations, letting us evaluate TransHands as a 2D-to-3D lifting module in isolation from 2D-detection errors. This isolates whether a body-motion encoder's semantic motion priors transfer to accurate 3D hand lifting, independent of detection noise. While this offers a clean controlled setting, TransHands remains inherently \textit{plug-and-play}, allowing the integration of any 2D tracker for real-world deployment (Section~\ref{subsec:zero_shot}).

\textbf{Transfer Protocol.} 
To evaluate our transfer strategy, we adopt a three-stage protocol that progressively relaxes the backbone from fully frozen to partially unfrozen, balancing prior preservation with domain adaptation.

\begin{itemize}
    \item \textbf{Stage 1: Geometric Alignment (S1).} 
    We initialize the backbone $\mathcal{E}_{body}$ with pre-trained body pose weights, keeping it \textit{frozen}. Only the topological adapter, the projection head, and the output decoder are optimized on Re:InterHand. Additionally, during this stage a self-supervised Lifted Masked Motion Modelling (L-MMM) objective (detailed in the supplementary materials) is activated to encourage robustness to occlusions by reconstructing full 3D poses from partially masked 2D inputs.
    
    \item \textbf{Stage 2-A: In-Domain Refinement (S2-A).}
    Starting from the \textbf{S1} checkpoint, we selectively unfreeze the last transformer blocks and optimize on Re:InterHand using a supervised \textit{Weighted MPJPE} loss ($\mathcal{L}_{sup}$), adapting the model to hand micro-dynamics while preserving the robust representations acquired during S1.

    \item \textbf{Stage 2-B: Multi-Dataset Generalization (S2-B).} 
    To address cross-domain generalization, we specialize the S1 model (aligned on Re:InterHand) on the target domains: \textbf{AssemblyHands} and \textbf{GigaHands}, using the same supervised objective as in S2-A. We employ a balanced sampling strategy~\cite{zhu2023motionbertunifiedperspectivelearning, pavlakos2023reconstructinghands3dtransformers} to ensure uniform exposure to all domains during this adaptation phase.
\end{itemize}

\textbf{Implementation Details.} 
All models are implemented in PyTorch and optimized using AdamW~\cite{loshchilov2019decoupledweightdecayregularization}.
We use a learning rate of $10^{-4}$ for Stage 1. For Stage 2 (both A and B), we apply a differential learning rate: $5 \times 10^{-6}$ for the unfrozen backbone layers to preserve pre-trained features, and $5 \times 10^{-5}$ for adapters. A Cosine Warmup scheduler~\cite{loshchilov2017sgdrstochasticgradientdescent, vaswani2023attention} stabilizes gradients at the start of each phase.  To facilitate reproducibility, we release our source code\footnote{Code available at: \url{https://github.com/picciolimilo/TransHands}\label{fn:code}}, including the training configurations for all backbones and stages.

\textbf{Ablation Studies.} Section~\ref{subsec:ablations} presents key ablations on our transfer strategy and its behaviour in low-data regimes, as well as on the key architectural components. Further analyses on the progressive training strategy and the use of the L-MMM objective in Stage S1 are provided in the supplementary materials.

\subsection{Main Results: Backbone Comparison}

Table~\ref{tab:indomain} reports adaptation efficiency and refinement on Re:InterHand, comparing frozen adaptation (S1) with partial unfreezing (S2-A).
To assess the framework's scalability, Table~\ref{tab:generalization} reports the cross-domain generalization performance; here we evaluate the models in two distinct scenarios: the zero-shot capability of the frozen body-centric prior (S1) and the performance of multi-dataset training (S2-B) on the same validation sets.
We report the Mean Per-Joint Position Error (\textbf{MPJPE})~\cite{martinez2017simple}, computed as the average Euclidean distance between predicted and ground-truth joints after root alignment, to measure absolute positional accuracy. As 2D lifting is scale-ambiguous, lifting predictions are normalized to a reference bone length, while the RGB baselines predict metric scale directly.
To assess the quality of the reconstructed structure independent of global factors, we also report the Procrustes-Aligned MPJPE (\textbf{PA-MPJPE}), which applies a rigid alignment (translation and rotation) to the prediction before error computation.
Additionally, qualitative visualizations illustrating the reconstruction quality across these domains are provided in the supplementary materials.

\input{tab/tab_inDomain}

\textbf{Transformer-based models dominate.}
Transformer-based architectures significantly outperform Graph Convolutional Networks in the final refinement stage. 
\textbf{MixSTE} stands as the best encoder in our benchmark, achieving the lowest error of \textbf{7.16~mm}. This represents a substantial improvement ($+21.34\%$) over its already strong frozen baseline.
\textbf{MotionBERT} follows closely, confirming the robustness of its dual-stream spatio-temporal attention mechanism with a remarkable gain of $+21.45\%$, reaching $7.47$~mm.

\textbf{Strongest Refinement Margin.}
A key observation from Table~\ref{tab:indomain} is that the highest adaptation gain does not strictly belong to the models with the largest capacity. \textbf{PoseFormerV2}, despite having significantly fewer parameters and a relatively higher initial error in S1 ($12.62$~mm) compared to the larger transformers in our benchmark, achieves the highest relative improvement in the benchmark ($+36.53\%$). This suggests that frequency-domain transformers possess a latent plasticity that, once unlocked in S2-A, allows for rapid convergence to the target domain geometry, successfully driving the error down to $8.01$~mm.

\textbf{Architecture Limits.}
\textbf{ST-GCN}, the smallest and only non-transformer model in the benchmark, appears to hit a capacity ceiling. Despite unlocking a parameter budget comparable to lightweight transformers during Stage 2 ($5.00$~M), it yields the highest S2-A error ($11.56$~mm) and the lowest overall gain ($+10.73\%$). This suggests that while partial refinement is highly effective for self-attention mechanisms, GCNs may struggle to handle significant domain shifts effectively through partial unfreezing alone. 

\input{tab/tab_crossDomain}

\textbf{Cross-Domain Generalization.}
Table~\ref{tab:generalization} reveals a compelling trade-off between intrinsic robustness and adaptation.
In the zero-shot setting (S1) on AssemblyHands, pure transformers exhibit superior robustness: \textbf{MixSTE} achieves the best initial alignment (\textbf{38.58~mm}), suggesting that its representation effectively encodes relative geometry, making it less sensitive to the global coordinate shifts typical of unseen domains.
Conversely, on GigaHands, all architectures yield high initial MPJPE values ($>115$ mm). However, their \textbf{PA-MPJPE} remains competitive ($\approx$ $24-30$~mm). This indicates that the high error stems from \textit{global rigid misalignment} (i.e., scale/rotation discrepancies between body and hand spaces) rather than structural distortion.
The multi-dataset adaptation (S2-B) resolves this misalignment, with high-capacity transformers leveraging their latent plasticity to recover the target distribution. \textbf{MixSTE} reaches \textbf{10.92~mm} on AssemblyHands, confirming that model capacity is a key driver for resolving geometric ambiguities. On GigaHands, the same adaptation drives all backbones from their initial $>115$~mm down to a few millimetres (e.g., \textbf{2.21~mm} for \textbf{MotionBERT}); this drop reflects the intrinsic nature of the domain rather than capacity alone, as GigaHands features more static, less articulated sequences with temporally smooth annotations, making it an easier target once the global misalignment is removed.

\subsection{Ablation Studies}
\label{subsec:ablations}

\subsubsection{The Value of Transfer Learning}
\label{subsec:ablation_transfer}

We hypothesize that the high performance of our framework relies on \textit{Transfer Learning}. Specifically, the adaptation of robust kinematic priors learned from full-body pose estimation, rather than solely learning from the target hand dataset.
To validate this, we test whether our pipeline achieves comparable results learning \textit{from scratch}.

\textbf{Setup.}
We compare two initialization strategies using exactly the same two-stage training protocol on the four motion encoder architectures:
\begin{enumerate}
    \item \textbf{Random Initialization (Baseline):} The encoder is initialized with standard random weights. This forces the model to learn 3D structures solely from the Re:InterHand dataset.
    \item \textbf{Body-Prior Initialization (Ours):} The encoder inherits pre-trained body weights, transferring learned spatial-temporal features and structural priors.
\end{enumerate}
To further evaluate the effectiveness of the transferred structural motion priors, we then assess TransHands in a low-data setting. We train the framework with PoseFormerV2 backbone using incrementally reduced fractions of the target dataset (from $100\%$ down to $10\%$) and compare it against the scratch baseline.

\textbf{Results.}
Table~\ref{tab:transfer_impact} highlights the performance gap across all stages.
During S1 (Frozen), the random baseline struggles with a higher error across all backbones, as a frozen random encoder produces noise that the projection head cannot map into valid poses. In contrast, our method starts with an initial semantic understanding, yielding an immediate MPJPE between 12.95 and 9.09 mm for all backbones.
When the encoder is allowed to learn during S2 (Partial Unfreeze), the random baseline recovers significantly, bringing the error down. However, it hits a performance ceiling. Our method achieves a superior final result compared to the baseline for all the backbones. PoseFormerV2 shows the strongest benefit from transfer learning, achieving a remarkable 35.7\% performance gain in stage S2. The data-efficiency analysis using this backbone further shows that TransHands matches the performance of the scratch baseline trained on the full dataset ($12.46$~mm) while using less than $25\%$ of the available annotations. In other words, TransHands reduces the annotation requirement by more than $75\%$ for PoseFormerV2, indicating that the transferred priors may effectively compensate for the limited availability of large-scale 3D hand datasets.

\begin{table}[!t]
\caption{\textbf{Transfer Learning Impact.} Comparison between random initialization (\textit{Scratch}) and our \textbf{TransHands} framework (\textit{Ours}) on Re:InterHand. The \textbf{Gain} column highlights the MPJPE reduction achieved at S2 through motion prior transfer.}
\label{tab:transfer_impact}
\centering
\begin{footnotesize}
\begin{sc}
\renewcommand{\arraystretch}{0.85}
\setlength{\tabcolsep}{3pt} 
\resizebox{\textwidth}{!}{%
\begin{tabular}{l cc cc cc cc c}
\toprule
\multirow{3}{*}{\textbf{Backbone}} & \multicolumn{4}{c}{\textbf{MPJPE} (mm) $\downarrow$} & \multicolumn{4}{c}{\textbf{PA-MPJPE} (mm) $\downarrow$} & \multirow{3}{*}{\textbf{S2 Gain (\%)}} \\
\cmidrule(lr){2-5} \cmidrule(lr){6-9}
& \multicolumn{2}{c}{\textbf{S1 (Frz)}} & \multicolumn{2}{c}{\textbf{S2 (Part)}} & \multicolumn{2}{c}{\textbf{S1 (Frz)}} & \multicolumn{2}{c}{\textbf{S2 (Part)}} & \\
\cmidrule(lr){2-3} \cmidrule(lr){4-5} \cmidrule(lr){6-7} \cmidrule(lr){8-9}
 & \textit{Scratch} & \textit{Ours} & \textit{Scratch} & \textit{Ours} & \textit{Scratch} & \textit{Ours} & \textit{Scratch} & \textit{Ours} & \\
\midrule
PoseFormerV2    & 19.67 & 12.62 & 12.46 & 8.01  & 10.97 & 7.88 & 8.58 & 5.88 & \textbf{35.7\%} \\
MotionBERT      & 10.76 & 9.51  & 8.40  & 7.47  & 8.00  & 6.81 & 6.47 & 5.62 & 11.1\% \\
ST-GCN          & 14.71 & 12.95 & 13.36 & 11.56 & 9.31  & 8.32 & 8.81 & 7.66 & 13.5\% \\
MixSTE          & 12.71 & \textbf{9.09}  & 8.88  & \textbf{7.16}  & 9.37  & \textbf{6.69} & 7.00 & \textbf{5.41} & 19.5\% \\
\bottomrule
\end{tabular}%
}
\end{sc}
\end{footnotesize}
\end{table}

\subsubsection{Architectural components}
We introduce a Neural ODE~\cite{chen2019neuralordinarydifferentialequations} adapter and a RetNet~\cite{sun2023retentivenetworksuccessortransformer} projection head to bridge the topological and dimensional gaps between hand and body sequences. Here, we isolate their contributions against standard baselines.

\textbf{Setup.} Using PoseFormerV2 and our two-stage training, we evaluate four configurations:
\begin{enumerate}
    \item \textbf{Baseline (MLP + Lin.):} Standard MLP adapter for input topology mapping and a Linear layer for output projection.
    \item \textbf{+ RetNet Proj.:} MLP adapter paired with RetNet projection to preserve temporal context.
    \item \textbf{+ ODE Adap.:} Continuous Neural ODE adapter paired with the baseline Linear projection.
    \item \textbf{Ours (Full):} The complete TransHands framework combining both the ODE adapter and the RetNet projection.
\end{enumerate}

\begin{table}[t]
\caption{\textbf{Architectural Components Ablation (PoseFormerV2).} Impact of replacing standard MLP and Linear layers with our proposed Neural ODE adapter and RetNet projection on Re:InterHand}
\label{tab:ablation_architecture}
\centering
\begin{footnotesize}
\begin{sc}
\renewcommand{\arraystretch}{0.85}
\setlength{\tabcolsep}{3pt} 
\resizebox{\textwidth}{!}{%
\begin{tabular}{l cc cc cc}
\toprule
\multirow{2}{*}{\textbf{Protocol}} & \multirow{2}{*}{\textbf{ODE Adap.}} & \multirow{2}{*}{\textbf{RetNet Proj.}} & \multicolumn{2}{c}{\textbf{MPJPE} $\downarrow$} & \multicolumn{2}{c}{\textbf{PA-MPJPE} $\downarrow$} \\
\cmidrule(lr){4-5} \cmidrule(lr){6-7}
 & & & \textbf{S1} & \textbf{S2-A} & \textbf{S1} & \textbf{S2-A} \\
\midrule
Baseline (MLP + Lin.) &            &            & 18.45 & 11.69 & 11.04 & 8.28 \\
+ RetNet Proj.        &            & \checkmark & 16.55 & 11.65 & 9.51 & 8.45 \\
+ ODE Adapter         & \checkmark &            & 13.72 & 9.47 & 8.81 & 6.97 \\
\textbf{Ours (Full)}  & \checkmark & \checkmark & \textbf{12.62} & \textbf{8.01} & \textbf{7.88} & \textbf{5.88} \\
\bottomrule
\end{tabular}%
}
\end{sc}
\end{footnotesize}
\end{table}

\textbf{Results.} Table~\ref{tab:ablation_architecture} isolates the contribution of each module. The baseline (MLP adapter + linear projection) reaches 11.69~mm at S2-A. Adding the RetNet projection alone yields a marginal change (11.65~mm), indicating that a stronger projection head provides little benefit without a matching input representation. The Neural ODE adapter, by contrast, drives the largest isolated drop (9.47~mm), as its continuous-depth formulation better models the highly non-linear mapping between hand and body joints. The two modules are complementary: combined, they reach the lowest error (\textbf{8.01~mm}), a further 1.46~mm over the ODE alone, showing that the RetNet projection becomes effective once the ODE aligns the input to the body manifold.

%% file: tab/tab_inDomain.tex
\begin{table}[t]
\caption{\textbf{Main Results In-Domain.} We report parameter efficiency and metric progression on the \textbf{Re:InterHand}~\cite{reinterhand} dataset for the four backbones (\cite{zhu2023motionbertunifiedperspectivelearning, zhang2022mixsteseq2seqmixedspatiotemporal, zhao2023poseformerv2exploringfrequencydomain, yan2018spatialtemporalgraphconvolutional}). \textbf{Total} denotes the backbone size, while \textbf{Tr-S1} and \textbf{Tr-S2} represent the trainable parameters in Stage 1 and 2 respectively. \textbf{S1} refers to frozen adaptation (alignment), and \textbf{S2-A} to partial unfreezing (refinement). \textbf{Gain} quantifies the relative MPJPE improvement from S1 to S2-A.}
\label{tab:indomain}
\centering
\begin{footnotesize}
\begin{sc}
\renewcommand{\arraystretch}{0.85}
\setlength{\tabcolsep}{3pt} 
\resizebox{\textwidth}{!}{%
\begin{tabular}{l ccc cc cc c}
\toprule
\multirow{2}{*}{\textbf{Backbone}} & \multicolumn{3}{c}{\textbf{Params (M)}} & \multicolumn{2}{c}{\textbf{MPJPE} $\downarrow$} & \multicolumn{2}{c}{\textbf{PA-MPJPE} $\downarrow$} & \multirow{2}{*}{\textbf{Gain (\%)}} \\
\cmidrule(lr){2-4} \cmidrule(lr){5-6} \cmidrule(lr){7-8}
& \textbf{Total} & \textbf{Tr-S1} & \textbf{Tr-S2} & \textbf{S1} & \textbf{S2-A} & \textbf{S1} & \textbf{S2-A} & \\
\midrule
MotionBERT     & 65.70 & 2.24 & 15.00 & 9.51          & 7.47          & 6.81          & 5.62          & $+$21.45 \\
MixSTE         & 36.02 & 2.24 & 10.79 & \textbf{9.09} & \textbf{7.16} & \textbf{6.69} & \textbf{5.41} & $+$21.34 \\
PoseFormerV2   & 16.62 & 2.24 & 5.89  & 12.62         & 8.01          & 7.88          & 5.88          & $+$36.53 \\
ST-GCN         & 6.61  & 4.10 & 5.00  & 12.95         & 11.56         & 8.32          & 7.66          & $+$10.73 \\
\bottomrule
\end{tabular}%
}
\end{sc}
\end{footnotesize}
\end{table}

%% file: tab/tab_crossDomain.tex
\begin{table}[t]
\caption{\textbf{Multi-Dataset Generalization.} We compare the zero-shot performance of the frozen backbone (\textbf{S1}) against the multi-dataset refined model (\textbf{S2-B}) on \textbf{AssemblyHands}~\cite{ohkawa2023assemblyhands} and \textbf{GigaHands}~\cite{fu2025gigahandsmassiveannotateddataset}.}
\label{tab:generalization}
\centering
\begin{footnotesize}
\begin{sc}
\renewcommand{\arraystretch}{0.85}
\setlength{\tabcolsep}{3pt} 
\resizebox{\textwidth}{!}{%
\begin{tabular}{l cccc cccc}
\toprule
& \multicolumn{4}{c}{\textbf{AssemblyHands}} & \multicolumn{4}{c}{\textbf{GigaHands}} \\
\cmidrule(lr){2-5} \cmidrule(lr){6-9}
& \multicolumn{2}{c}{\textbf{MPJPE} $\downarrow$} & \multicolumn{2}{c}{\textbf{PA-MPJPE} $\downarrow$} & \multicolumn{2}{c}{\textbf{MPJPE} $\downarrow$} & \multicolumn{2}{c}{\textbf{PA-MPJPE} $\downarrow$} \\
\cmidrule(lr){2-3} \cmidrule(lr){4-5} \cmidrule(lr){6-7} \cmidrule(lr){8-9}
\textbf{Backbone} & \textbf{S1} & \textbf{S2-B} & \textbf{S1} & \textbf{S2-B} & \textbf{S1} & \textbf{S2-B} & \textbf{S1} & \textbf{S2-B} \\
\midrule
MotionBERT     & 42.87 & 11.73 & 22.06 & 6.96 & 139.00 & \textbf{2.21} & 29.80 & \textbf{1.89} \\
MixSTE         & \textbf{38.58} & \textbf{10.92} & \textbf{21.71} & \textbf{6.47} & 121.20 & 2.36 & 28.27 & 2.03 \\
PoseFormerV2   & 39.70 & 13.05 & 21.93 & 7.34 & \textbf{115.90} & 2.82 & 27.68 & 2.37 \\
ST-GCN         & 54.80 & 20.66 & 24.15 & 11.34 & 137.02 & 6.12 & \textbf{26.99} & 4.38 \\
\bottomrule
\end{tabular}%
}
\end{sc}
\end{footnotesize}
\end{table}

%% file: src/sec_robustness.tex
\section{Robustness and Generalization in Real-World Contexts}
\label{sec:robustness}

In real-world deployment, hand pose estimation models must remain reliable under severe domain shift, handle noisy 2D detector inputs from unconstrained environments, and support downstream tasks such as gesture recognition under privacy-preserving constraints, that may preclude the retention of raw RGB data. This section compares TransHands along all these axes with respect to the current state of the art.
First, we evaluate cross-site generalization under significant domain shift, testing the framework's resilience to realistic 2D detector noise without any target-domain supervision (Section~\ref{subsec:zero_shot}). Second, we validate that our compact, skeleton-only representation encodes semantically rich and transferable motion priors through a downstream gesture recognition task across both exo-centric and egocentric vision (Section~\ref{subsec:gesture_recognition}).

\subsection{RGB-to-3D Hand Pose Estimation}
\label{subsec:zero_shot}

We evaluate robustness under domain shift, from Re:InterHand to the unseen AssemblyHands domain, using an off-the-shelf 2D keypoints detector as input to the TransHands lifting module.

\textbf{Zero-Shot Generalization.}
We evaluate our best S2-A models (MixSTE and MotionBERT backbones) directly on the AssemblyHands validation set, without fine-tuning; for each hand we aggregate the four ego camera views, taking the prediction from the first view that observes it. Unlike the in-domain adaptation of Table~\ref{tab:generalization}, the S2-A model (trained only on Re:InterHand) here receives noisy keypoints detected using MediaPipe~\cite{lugaresi2019mediapipeframeworkbuildingperception} rather than ground-truth 2D annotations. Table~\ref{tab:zero_shot_results} shows that our approach reaches an MPJPE of \textbf{92.97~mm} and a PA-MPJPE of $23.82$~mm with the MotionBERT backbone ($93.44$~mm / $24.28$~mm with MixSTE).
\input{tab/tab_robustness}
\input{tab/tab_gesture}

Despite relying only on sparse 2D keypoints from an off-the-shelf detector, our model improves on the reported MPJPE of the RGB baselines ArcticNet-SF~\cite{fan2023arcticdatasetdexterousbimanual} ($110.76$~mm). On a root-aligned basis, TransHands ($92.97$~mm) is competitive with the dedicated egocentric method V-HPOT~\cite{mucha2026egocentric3dhandpose} ($92.09$~mm). This compares different regimes rather than claiming superiority: V-HPOT is an end-to-end RGB model that recovers metric depth and adapts at test time, whereas TransHands lifts noisy off-the-shelf 2D keypoints zero-shot, reaching comparable accuracy without any target-domain adaptation. Moreover, this result improves the performance, on the same 2D keypoints, of MediaPipe~3D which applies the GHUM model~\cite{xu2020ghum} to perform 2D-to-3D uplifting. Mediapipe~3D reaches $130.77$~mm MPJPE; our pipeline recovers about $38$~mm, denoting a better understanding of the 3D hand geometry compared to the uplifting strategy employed by MediaPipe. We also compare with SMPLer-X~\cite{cai2024smplerxscalingexpressivehuman}, a SOTA whole-body mesh model, which motivates a dedicated hand-pose task in egocentric settings where full-body visibility is limited. Despite this limitation, SMPLer-X attains a lower PA-MPJPE ($19.40$~mm): its parametric SMPL-X prior enforces anatomically valid hand shapes and is, by construction, immune to the depth-sign ambiguity of monocular lifting. Our pipeline, however, achieves a lower MPJPE (92.97 mm vs. 98.20 mm) with approximately one-tenth as many parameters (67.7M, including the MediaPipe 2D hand detector, vs. 703.58M for the end-to-end SMPLer-X pipeline). This parameter comparison is contextual, as SMPLer-X targets the much broader task of full-body mesh recovery.

\subsection{Downstream Utility: Gesture Recognition}
\label{subsec:gesture_recognition}

To evaluate the generalizability and semantic richness of the spatial-temporal representations learned by our framework, we assess TransHands on a downstream action recognition task. Specifically, we extract the latent embeddings from the S2-B backbones and use them to classify hand gestures.

\textbf{Experimental Setup.} 
We freeze the pre-trained encoder and projection of TransHands, training only 
the input adapter, to strictly evaluate the quality of the pre-computed features.  
The latent representations are then processed by a Temporal Convolutional Network (TCN)~\cite{bai2018empiricalevaluationgenericconvolutional} classification head.
The network is trained using a Gesture Classification Loss combining Focal Loss~\cite{lin2018focallossdenseobject} and Label Smoothing~\cite{szegedy2015rethinkinginceptionarchitecturecomputer} to handle class imbalance. Optimization is performed via AdamW with a Warmup-Cosine learning rate schedule.

\textbf{Datasets and Domains.} 
We evaluate the representations on two distinct domains: \textbf{Jester}~\cite{materzynska2019jester}, a large-scale dataset of human hand gestures recorded from a fixed, third-person webcam (static domain), and \textbf{EgoGesture}~\cite{zhang2018egogesture}, a challenging dataset of egocentric gestures recorded from wearable cameras, featuring severe perspective distortions and head motion (dynamic first-person domain).

\textbf{Results.} 
Table~\ref{tab:gesture_results} summarizes the Top-1 and Top-5 accuracy of the frozen representations extracted from MotionBERT and MixSTE. 
Both models demonstrate cross-domain transferability. Reaching 88\% on the Jester dataset and $\sim$88\% on the EgoGesture dataset 
with a \textit{frozen} lifting encoder confirms that our learned motion priors encode high-level semantic dynamics.

\textbf{Comparison with Literature.}
To contextualize these results, full-RGB models on \textbf{Jester} (e.g., TSM~\cite{lin2019tsmtemporalshiftmodule}) exceed 95\% accuracy by exploiting contextual cues at a high computational cost. In contrast to the pose-only method of Schl\"usener et al.~\cite{schluesener2022fastlearningdynamichand} (81.2\% on a 10-class subset), TransHands achieves over 88\% on the full 27-class vocabulary. On \textbf{EgoGesture}, our model ($\sim$88\%) outperforms RGB baselines like VGG16+LSTM~\cite{donahue2016longtermrecurrentconvolutionalnetworks} (74.7\%) and C3D~\cite{tran2015learningspatiotemporalfeatures3d} (86.4\%). While complex RGB-Depth ensembles~\cite{cao2017egocentric} reach 92.2\%, TransHands provides a highly efficient and privacy-preserving alternative.

\subsection{Limitations and Future Work}
Our results establish the benefit of body-motion transfer within our multi-dataset and multi-backbone framework. Further comparisons with hand-specific 2D-to-3D lifting methods are left open for future work. In addition, our evaluation relies  mostly on ground-truth 2D keypoints, real-world accuracy remains bounded by the 2D detector (Section \ref{subsec:zero_shot}). Further integration between 2D estimation and 3D-uplifting through TransHands is left for future analysis.

%% file: tab/tab_robustness.tex
\begin{table}[ht]
\caption{\textbf{Zero-Shot Robustness Analysis on AssemblyHands.} Evaluation on the unseen AssemblyHands domain. \textit{Lift} denotes methods that estimate 2D keypoints with an off-the-shelf detector and then uplift them to 3D (our pipeline and MediaPipe~3D, which share the same 2D input); \textit{RGB} denotes end-to-end methods that estimate 3D directly from the image.}
\label{tab:zero_shot_results}
\centering
\begin{footnotesize}
\begin{sc}
\renewcommand{\arraystretch}{0.95}
\setlength{\tabcolsep}{6pt}
\resizebox{\columnwidth}{!}{%
\begin{tabular}{l c c c}
\toprule
\textbf{Method} &
\textbf{Type} &
\textbf{MPJPE} $\downarrow$ &
\textbf{PA-MPJPE} $\downarrow$ \\
\midrule
ArcticNet-SF~\cite{fan2023arcticdatasetdexterousbimanual} & RGB & 110.76$^\dag$ & -- \\
V-HPOT~\cite{mucha2026egocentric3dhandpose}               & RGB & \textbf{92.09}$^\ddag$ & -- \\
SMPLer-X~\cite{cai2024smplerxscalingexpressivehuman}      & RGB & 98.20 & \textbf{19.40} \\
MediaPipe 3D~\cite{lugaresi2019mediapipeframeworkbuildingperception} & Lift & 130.77 & 29.38 \\
\textbf{Ours (MotionBERT)} & Lift & 92.97 & 23.82 \\
\textbf{Ours (MixSTE)}     & Lift & 93.44 & 24.28 \\
\bottomrule
\\[-8pt]
\multicolumn{4}{r}{\footnotesize \upshape $^\dag$ As reported in \cite{prakash20243dhandposeestimation}; $^\ddag$ As reported in \cite{mucha2026egocentric3dhandpose}.} \\
\end{tabular}%
}
\end{sc}
\end{footnotesize}
\end{table}

%% file: tab/tab_gesture.tex
\begin{table}[t]
\caption{\textbf{Downstream Gesture Recognition.} Evaluation of the frozen S2-B representations on Jester and EgoGesture datasets.}
\label{tab:gesture_results}
\centering
\begin{footnotesize}
\begin{sc}
\renewcommand{\arraystretch}{0.85}
\setlength{\tabcolsep}{3pt} 
\resizebox{\textwidth}{!}{%
\begin{tabular}{l c c c c}
\toprule
\multirow{2}{*}{\textbf{Backbone}} & \multicolumn{2}{c}{\textbf{Jester}} & \multicolumn{2}{c}{\textbf{EgoGesture}} \\
\cmidrule(lr){2-3} \cmidrule(lr){4-5}
 & \textbf{Top-1} (\%) $\uparrow$ & \textbf{Top-5} (\%) $\uparrow$ & \textbf{Top-1} (\%) $\uparrow$ & \textbf{Top-5} (\%) $\uparrow$ \\
\midrule
MotionBERT & 88.44 & \textbf{95.80} & \textbf{88.29} & 97.36 \\
MixSTE     & \textbf{88.72} & 95.76 & 88.08 & \textbf{97.53} \\
\bottomrule
\end{tabular}%
}
\end{sc}
\end{footnotesize}
\end{table}

%% file: src/sec_conclusion.tex
\section{Conclusion}
\label{sec:conclusion}
We presented a framework that repurposes body-motion encoders for 3D hand pose lifting from monocular 2D inputs, decoupling topological adaptation from temporal modeling to enable cross-topology transfer with minimal architectural change. Across multiple encoders, body-derived priors provide strong inductive biases for data-efficient and robust hand pose lifting, highlighting the potential of motion representation reuse across articulated domains and its relevance for downstream applications in real-world contexts.